\pdfoutput=1
\documentclass[letterpaper]{article} % DO NOT CHANGE THIS
\usepackage{aaai23}  % DO NOT CHANGE THIS
\usepackage{times}  % DO NOT CHANGE THIS
\usepackage{helvet}  % DO NOT CHANGE THIS
\pdfoutput=1
\usepackage{courier}  % DO NOT CHANGE THIS
\usepackage[hyphens]{url}  % DO NOT CHANGE THIS
\usepackage{graphicx} % DO NOT CHANGE THIS
\usepackage{natbib}  % DO NOT CHANGE THIS AND DO NOT ADD ANY OPTIONS TO IT
\usepackage{caption} % DO NOT CHANGE THIS AND DO NOT ADD ANY OPTIONS TO IT
\usepackage{algorithm}
\usepackage{algorithmic}
\usepackage[switch]{lineno}
\usepackage{adjustbox}
\usepackage{colortbl}
\usepackage{multirow}
\usepackage{makecell}
\makeatletter
  \newcommand\figcaption{\def\@captype{figure}\caption}
  \newcommand\tabcaption{\def\@captype{table}\caption}
\makeatother

\usepackage{booktabs}
\usepackage{amsmath}
\usepackage[table]{xcolor}
\usepackage{amssymb}
\usepackage{newfloat}
\usepackage{listings}
\DeclareCaptionStyle{ruled}{labelfont=normalfont,labelsep=colon,strut=off} % DO NOT CHANGE THIS
\floatstyle{ruled}
\newfloat{listing}{tb}{lst}{}
\floatname{listing}{Listing}
\nocopyright %-- Your paper will not be published if you use this command
\title{CALIP: Zero-Shot Enhancement of CLIP with Parameter-free Attention}
\vspace{0.2cm}
\author {
    Ziyu Guo\textsuperscript{\rm1*},
    Renrui Zhang\textsuperscript{\rm1,2*},
    Longtian Qiu\textsuperscript{\rm3*},
    Xianzheng Ma\textsuperscript{\rm2},\\
    Xupeng Miao\textsuperscript{\rm1},
    Xuming He\textsuperscript{\rm3},
    Bin Cui\textsuperscript{\rm1}
    \vspace{0.1cm}
}

\affiliations {
    \textsuperscript{\rm 1} School of CS and Key Lab of HCST, Peking University\\
    \textsuperscript{\rm 2} Shanghai AI Laboratory\ \ \ 
    \textsuperscript{\rm 3} ShanghaiTech University\\
    \vspace{0.1cm}
    \{guo.ziyu, bin.cui\}@pku.edu.cn, zhangrenrui@pjlab.org.cn
    \vspace{0.1cm}
}

\begin{document}
% \linenumbers
\maketitle

\begin{abstract}
% \vspace{0.5cm}
Contrastive Language-Image Pre-training (CLIP) has been shown to learn visual representations with promising zero-shot performance. To further improve its downstream accuracy, existing works propose additional learnable modules upon CLIP and fine-tune them by few-shot training sets. However, the resulting extra training cost and data requirement severely hinder the efficiency for model deployment and knowledge transfer. In this paper, we introduce a free-lunch enhancement method, \textit{\textbf{CALIP}}, to boost \textit{\textbf{CLIP}}'s zero-shot performance via a parameter-free \textit{\textbf{A}}ttention module. Specifically, we guide visual and textual representations to interact with each other and explore cross-modal informative features via attention. As the pre-training has largely reduced the embedding distances between two modalities, we discard all learnable parameters in the attention and bidirectionally update the multi-modal features, enabling the whole process to be parameter-free and training-free. In this way, the images are blended with textual-aware signals and the text representations become visual-guided for better adaptive zero-shot alignment. We evaluate CALIP on various benchmarks of 14 datasets for both 2D image and 3D point cloud few-shot classification, showing consistent zero-shot performance improvement over CLIP. Based on that, we further insert a small number of linear layers in CALIP's attention module and verify our robustness under the few-shot settings, which also achieves leading performance compared to existing methods. Those extensive experiments demonstrate the superiority of our approach for efficient enhancement of CLIP. Code is available at \url{https://github.com/ZiyuGuo99/CALIP}.
\end{abstract}

\begin{figure}[h]
\vspace{0.4cm}
    \includegraphics[width=0.95\linewidth]{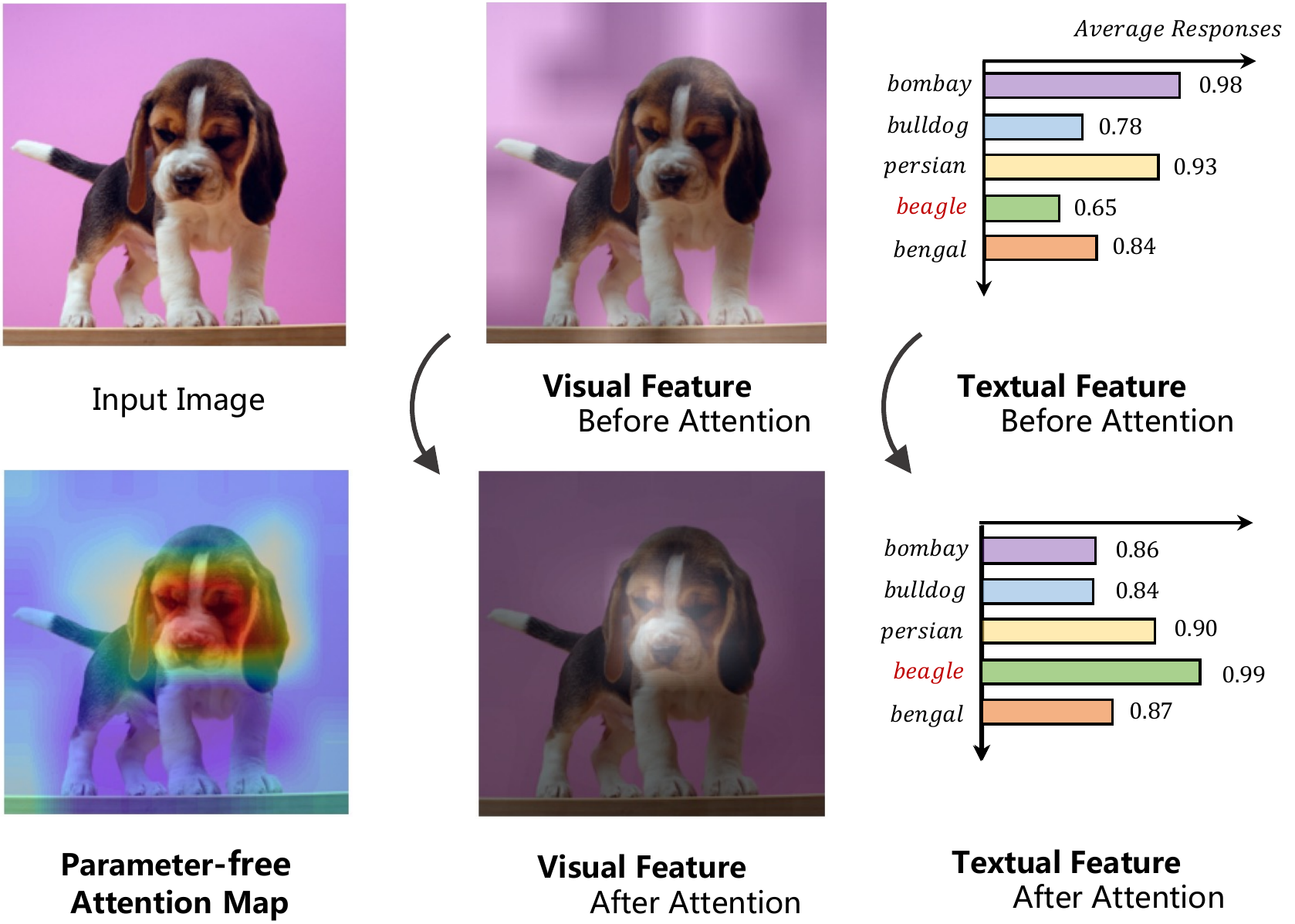}
  \caption{Visualization of Parameter-free Attention and the Interacted Features. Without any parameters, CALIP's cross-modal attention map (Left-Bottom) shows favorable weight distributions over the main objects, which well updates both visual and textual features: pixels within objects of ground-truth labels are enhanced and the corresponding category features in red are strengthened.}
  \label{calip_teaser}
%   \vspace{0.5cm}
\end{figure}

\section{Introduction}
With the advance of learning theories and network architectures, supervised methods under a close-set assumption have achieved extraordinary results over a wide range of vision tasks, such as image classification~\cite{resent,krizhevsky2012imagenet,parmar2018image,mao2021dual}, object detection~\cite{ren2015faster,carion2020end,zheng2020end,chen2017multi}, and point cloud understanding~\cite{qi2017pointnet,qi2017pointnet++}. Despite their success in those specific scenarios, they often lack the ability to attain general visual representations, which harms their transferability to open-set applications. Alternatively, based on exploiting the wide coverage of languages, Contrastive Language-Image Pre-training (CLIP)~\cite{clip} proposes to conduct visual learning contrastively with descriptive natural language data. Pre-trained by large-scale image-text pairs, CLIP extracts both features of input images and texts by independent encoders, and aligns the paired ones within the same embedding space. 
% Such cross-modal alignment endows CLIP with the capability of recognizing new visual concepts on downstream tasks. 
On downstream tasks, given a new dataset with images of ``unseen'' classes, CLIP constructs the textual inputs by the category names and converts the original classification task into a image-text matching problem. As such, CLIP is able to achieve zero-shot recognition in open-vocabulary settings and obtains promising performance on various benchmarks.

To further improve the downstream performance of CLIP, existing works introduce different fine-tuning methods for the few-shot classification. Inspired by prompt tuning~\cite{li2021prefix} and adapters~\cite{houlsby2019parameter} in natural language processing, Context Optimization (CoOp)~\cite{coop}, CLIP-Adapter~\cite{adapter} and Tip-Adapter~\cite{zhang2021tip} freeze CLIP's pre-trained weights and adopt learnable prompts or lightweight adapters to tune the textual and visual features.
% best-fitted textual inputs  other than the original hand-crafted ones. CLIP-Adapter~\cite{adapter} appends a lightweight adapter module~\cite{houlsby2019parameter} upon the visual encoder and updates CLIP with the adapted image features to improve the performance. Tip-Adapter~\cite{zhang2021tip} constructs a cache model from the few-shot training set and greatly reduces the implementation costs for CLIP's adaption.
% PointCLIP~\cite{zhang2021pointclip}, DenseCLIP~\cite{rao2021denseclip}, ActionCLIP~\cite{wang2021actionclip} also enhance CLIP's recognition ability for other domains aided by few-shot fine-tuning. 
Despite the performance improvement, all existing methods with task-specific designs contain learnable parameters and rely on additional training phase with few-shot labeled data. This leads to extra resource cost and largely hinders CLIP's inherent advantage for efficient zero-shot knowledge transfer. As an example, existing methods are required to fine-tune CLIP separately for different downstream tasks, and deploy multiple model copies for different applications. Therefore, we ask the question: \textbf{Can we adapt CLIP by a more efficient and general method without additional few-shot data or training?}
% \vspace{0.1cm}

\begin{figure*}[t!]
  \centering
    \includegraphics[width=0.9\textwidth]{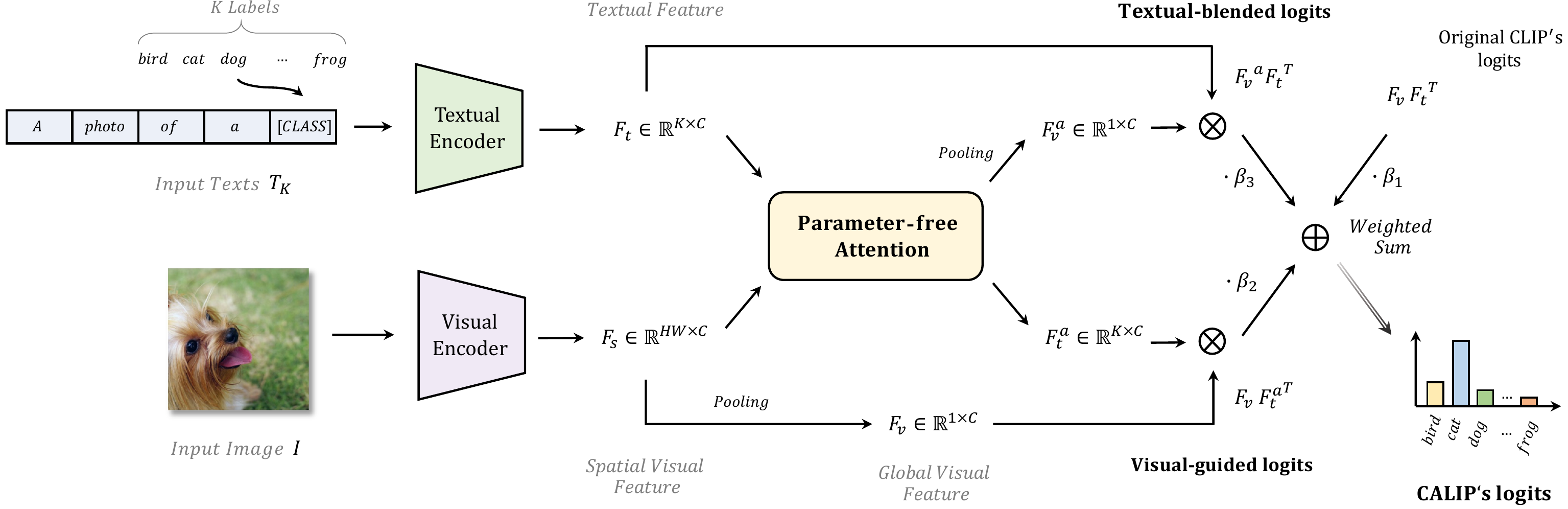}
   \caption{The Pipeline of CALIP. We introduce a parameter-free attention module for zero-shot enhancement of CLIP and require no extra data or training for downstream tasks. CALIP utilizes pre-trained encoders to extract spatial visual feature of the input image and $K$-category textual feature. Then, the proposed attention module updates their representations via cross-modal interactions and outputs the final zero-shot prediction by weighted summation of three classification logits.}
    \label{calip_p1}
\end{figure*}

To tackle this issue, we propose \textbf{CALIP}, which equips CLIP with a parameter-free attention module to conduct cross-modal interactions and avoid the need for extra downstream data or training, as shown in Figure~\ref{calip_p1}. Before the CLIP outputting the final global feature of an image, we utilize its intermediate feature map, which preserves more fine-grained semantic information and contextual characteristics of the image. Then, we conduct a parameter-free cross-modal attention between the spatial visual feature and the textual feature, containing no learnable parameter. Different from traditional attention mechanism, our design consists of two key modifications, which are non-parametric and bidirectional. For the former, as the features of CLIP's two modalities have been well aligned during the contrastive pre-training, we are able to simply omit the linear layers within the attention, which were supposed to project the features into queries, keys and values. Therefore, their attention map can be directly calculated by matrix multiplication between features. For the latter, as there is no discrimination for queries, keys or values, we can simultaneously update both visual and textual features via the only attention map.
% Specifically, traditional attention mechanism takes as input two terms, and applies linear layers to project the first one into queries and the other into keys and values. Such projection is to unify the embedding space for proper similarities calculation between queries and keys. Based on the similarity matrix, the first input term is updated by aggregating features from the other though weighted summation of values and post-processing via a back-projection linear layer. However for ours, considering that the features of two modalities have been well aligned during the contrastive pre-training, we could simply omit the projection layers and obtain the matrix by direct calculation between CLIP-encoded visual and textual features. As there is no discrimination for queries, keys or values, we simultaneously update both features via the same similarity matrix. 
With this attention mechanism, the visual feature is guided by category semantics from the texts, which converts their per-pixel features to be more distinctive for recognition. 
Correspondingly, the text counterpart adaptively explores features from informative regions on the image and becomes visual-aware and image-conditional, instead of remaining the same for the entire dataset. The visualization in Figure~\ref{calip_teaser} demonstrates the effectiveness of our parametric-free attention. Finally, the zero-shot prediction of CALIP is obtained by matching between the visual and textual features after our proposed cross-modal interactions.

The whole process of CALIP is zero-shot, training-free and universal for various downstream tasks. We implement and evaluate CALIP on 14 datasets including zero-shot 2D image and 3D point cloud classification to illustrate its effectiveness. For some benchmarks, zero-shot CALIP without training even surpasses some prior methods after few-shot fine-tuning. On top of that, to fully unleash the power of cross-modal attention, we further add a small number of linear layers in the attention module and upgrade the parameter-free attention into a parametric version, named \textbf{CALIP-FS}. Under the few-shot fine-tuning, CALIP-FS achieves leading performance among all existing methods, which demonstrates the superiority of our proposed attention framework. The main contributions of CALIP are as follows:
% \vspace{-25pt}
\begin{itemize}
    \item To our best knowledge, CALIP is the \textbf{first work} to conduct zero-shot enhancement over CLIP for downstream tasks without few-shot data or additional training.
    \item We design a parameter-free attention for cross-modal interactions upon CLIP to effectively exchange image-text informative features for better alignment.
    \item The parametric version, CALIP-FS with learnbale cross-modal attention modules, also achieves competitive performance among all existing few-shot methods.
    % \item We conduct extensive experiments on 14 well-known datasets for 2D image and 3D point cloud classification to demonstrate the advantages and robustness of our method.
\end{itemize}

% \paragraph{\textbf{Vision-Language Models.}} Exploring the interaction between vision and language has motivated various multi-modal tasks, including image captioning~\cite{xu2015show}, visual question answering~\cite{antol2015vqa}, visual grounding~\cite{chen2019see} and so on. The two modalities together can provide rich and complementary information, which benefits the learning of both representations. Recently, Contrastive Language-Image Pre-training (CLIP)~\cite{clip} has shown great potential for obtaining generic visual representations by contrastive learning and achieves inspirational downstream performance. The follow-up works further use noisy data for better scaling ability~\cite{jia2021scaling} or introduce the data-efficient scheme with extra supervisions~\cite{li2021supervision}. However, all their designs aim to improve its learning behaviors during the pre-training phase and such re-pre-training of CLIP consumes quite expensive time and computation resources. In contrast, our CALIP is proposed to directly boost the pre-trained CLIP by a parameter-free attention module without any data or training, which is both efficient and effective.

% Early researches leverage object detection model to extract image feature to explore the interaction between text and image\cite{LXMERT} \cite{ViLBERT}\cite{gao2019dynamic}. Inspired by the success of pre-training models BERT\cite{BERT}, UNITER\cite{UNITER} and Oscar\cite{Oscar} use attention architecture further improve the performance of vision-language tasks. 
% At present, the breakthrough in vision-language learning, particularly CLIP\cite{clip} is driven by the noisy large-scale datasets available in the Internet, which are 400 million image-text pairs. The contrastive learning strategy on a huge amount of image-text pairs are employed by CLIP\cite{clip} and show impressive transferable ability on downstream tasks.
\section{RELATED WORK}

\paragraph{\textbf{Downstream Adaption of CLIP.}}
As a breakthrough in vision-language learning, CLIP~\cite{clip} has shown great potential for obtaining generic visual representations by contrastive pre-training. Based on the superior transferable ability, the problem of effectively adapting CLIP to downstream tasks has been widely studied. Given few-shot training data, CoOp~\cite{coop} proposes the learnable prompts for textual inputs inspired by prompt learning~\cite{li2021prefix}, and VT-CLIP~\cite{zhang2021vt} introduces visual-guided texts for better vision-language alignment. Referring to adapters~\cite{houlsby2019parameter}, CLIP-Adapter~\cite{adapter} appends a lightweight adapter module to produce adapted multi-modal features. Tip-Adapter~\cite{zhang2021tip} and CoMo~\cite{zhang2022collaboration} greatly reduce its training cost by constructing a key-value cache model. Besides 2D, PointCLIP V1~\cite{zhang2021pointclip} and V2~\cite{zhu2022pointclip} extend CLIP into 3D data understanding by projecting point clouds into multi-view depth maps. Other works also apply CLIP for object detection~\cite{du2022learning}, semantic segmentation~\cite{rao2021denseclip}, depth estimation~\cite{zhang2022can} and video analysis~\cite{lin2022frozen}. However, the existing downstream adaption of CLIP demands extra training data and the resources for fine-tuning, which weakens CLIP's core advantage of efficient zero-shot recognition. In this paper, we explore CALIP to enhance CLIP's downstream performance under zero-shot settings by interacting its two modalities with no parameter. In addition, our approach can be utilized for both 2D and 3D domains and is also well-performed when few-shot data are available, indicating great generalization ability.

\paragraph{\textbf{Attention in Vision.}}
Attention mechanism is introduced by~\cite{vaswani} in natural language processing, which calculates pair-wise similarities of input tokens and is advantageous to extract long-range dependencies. In computer vision tasks, ViTs~\cite{dosovitskiy2021vit,mao2021dual} view image patches as tokens and utilize self-attention to model their long-range dependencies. DETRs~\cite{carion2020end,zheng2020end} convert object detection into set prediction problems by bipartite matching and apply cross-attention to decode learnable queries for parallel outputs. Besides, attention-based networks also largely benefit image segmentation~\cite{xie2021segformer}, video classification~\cite{li2022uniformer}, visual question answering~\cite{vqa}, object tracking~\cite{meinhardt2021trackformer}, 3D object detection~\cite{zhang2022monodetr} and point cloud classification~\cite{guo2021pct,zhang2022point}.
Different from all existing methods, whose attention modules are training-based and contain learnable parameters, we design a parameter-free attention module to achieve attention calculation by pure matrix multiplications and bidirectionally interact the multi-modal features. In this way, CALIP could not only profit by the powerful attention, but also transfer to downstream tasks under zero-shot circumstances. 
% Moreover, our attention is bidirectional for both modalities.

% The basic module is self-attention layers composed by multi-head scaled dot-product attention and feed forward network which is vital for extracting characteristics of relationships between words at multiple levels. One of the main advantages of attention methods is their non-local computations and perfect global memory, which makes them more suitable than RNNs on dealing with sequential data
% \cite{huang2017instance}
% \cite{huang2015bidirectional}
% .Recently, cross attention mechanism is designed to fuse inputs from different feature data in the embedding space which has been widely applied to various visual tasks, including object detection\cite{detr}
% \cite{ji2020casnet}
% \cite{lee2018stacked}
% and image classification\cite{hou19}
% \cite{chen2021crossvit}
% \cite{liu2021multiscale}
% which demonstrate the potential of cross attention mechanism. SMCA\cite{gao2021fast} proposed a Modulated Attention mechanism which can achieve fast-convergence on object detection. Motivated by the success of previous works, we propose a non-parametric attention where all learnable linear layers are removed and non-parametric attention is directly computed by pure matrix multiplication.

\section{Method}
In this section, we first revisit CLIP for zero-shot recognition as the preliminary. Then we present the details of our zero-shot CALIP with parameter-free attention, followed by the parametric version, CALIP-FS.

\subsection{Preliminary of CLIP}
\label{clip_m}
CLIP utilizes 400 million image-text pairs for contrastive pre-training in an unsupervised way, obtaining the ability to match ``unseen'' images with their corresponding categories. To extract features of both modalities, CLIP has two independent encoders: a ResNet~\cite{resent} or vision transformer (ViT)~\cite{dosovitskiy2021vit} for visual encoding, and a 12-layer transformer~\cite{vaswani} for textual encoding, denoted as $\mathrm{VisEnc(\cdot)}$ and $\mathrm{TexEnc(\cdot)}$, respectively. For the downstream dataset with $K$ categories, $\{ C_1, C_2\dots, C_K \}$, CLIP places all category names into the [CLASS] token of a pre-defined textual template, e.g., ``a photo of a [CLASS]'', constructing $K$ textual inputs $\mathrm{T_K}$. Then, their textual features are extracted as $F_t\in R^{K\times C}$, whose $i$-th row vector, $i=1,\dots,K$, represents the encoded knowledge of category $C_i$. For every input image $\mathrm{I}$ to be recognized, CLIP extracts its spatial feature map $F_{s}\in R^{H\times{W\times{C}}}$ and obtains the global visual representation $F_{v}\in R^{1\times{C}}$ by pooling operation. Finally, features from both encoders are matched via cosine similarities to produce the classification $logits\in R^{1\times{K}}$. The whole process is as
\begin{align}
    &F_t = \mathrm{TexEnc}(\mathrm{T_K}),\\
    &F_v = \mathrm{Pooling}(F_{s}),\ \ F_{s} = \mathrm{VisEnc}(\mathrm{I}),\\
    &logits = F_v F_t^T,
\end{align}
where we assume $F_v$ and $F_t^T$ are L2-normalized features and their matrix multiplication is equal to cosine similarities calculation. $logits$ denote the probabilities for all $K$ categories and CLIP outputs the maximum one as the prediction. 

\begin{figure*}[t!]
  \centering
    \includegraphics[width=0.9\textwidth]{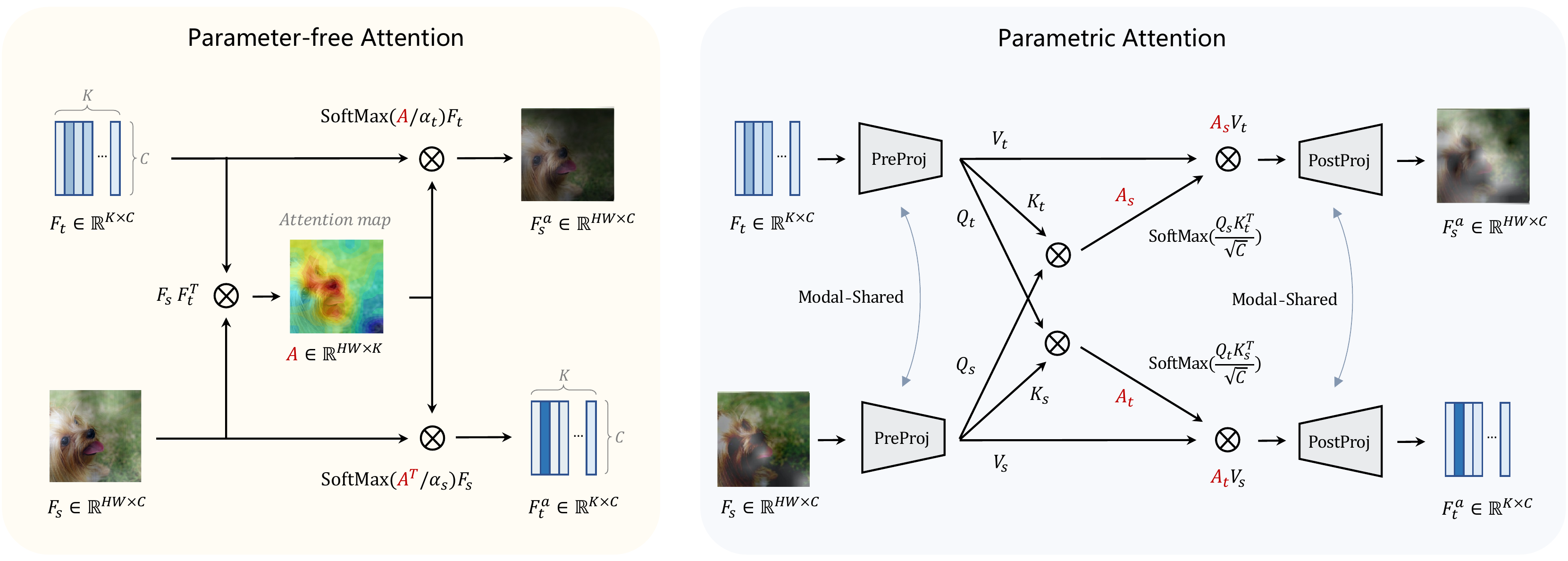}
   \figcaption{Structures of Parameter-free (Left) and Parametric Attention (Right). Parameter-free attention directly obtains the cross-modal attention map $A$ by matrix multiplication and bidirectionally updates two features for zero-shot classification. Parametric attention is equipped with both pre-projection and post-projection layers for better few-shot performance.}
    \label{calip_p2}
\end{figure*}

\subsection{CALIP with Parameter-free Attention}
\label{calip}
\paragraph{\textbf{Motivation.}}
While CLIP achieves promising results on zero-shot open-vocabulary recognition, which is concise and efficient, it still has room for improvement. We observe that the two modalities are totally isolated during encoding and there is no bridge for inter-modal information flow before the final matching. In addition, the spatial structures of images in $F_s$ are largely left out by the pooling operation, which might harm the fine-grained visual understanding. More importantly, we aim to inherit the great strength of CLIP's zero-shot capacity for training-free transfer learning, which requires no downstream data. Therefore, we propose our parameter-free attention module (CALIP) to not only fulfill the cross-modal interactions, but also achieve the goal to conduct zero-shot enhancement over CLIP.

\paragraph{\textbf{Design Details.}}
After CLIP's encoding of two modalities, we utilize the intermediate spatial visual feature $F_s \in R^{H\times{W\times{C}}}$ and textual feature $F_t\in R^{K\times{C}}$ for interactions. We reshape $F_s$ into a 1D vector sequence, $F_s \in R^{HW\times{C}}$ and obtain their attention weights directly by matrix multiplication without any projection,
\begin{align}
    A = F_s F_t^T\ \in R^{HW\times K},
\end{align}
where $A$ denotes the cross-modal attention map. Each element of $A$ represents the attention weight, namely, the feature similarity between a category and one image pixel/site. Based on $A$, we bidirectionally update both textual and visual features as follows
\begin{align}
    &F_s^a = \mathrm{SoftMax}(A/\alpha_t)F_t,\\
    &F_t^a = \mathrm{SoftMax}(A^T/\alpha_s)F_s,
\end{align}
where $\alpha_t$ and $\alpha_s$ modulate the attention magnitude for textual and visual modalities, respectively. Weighted by the attention scores representing similarity, two modalities both aggregate informative features from each other as visualized in Figure~\ref{calip_teaser}. For texts, as $F_t$ encodes $K$-category knowledge, the signals of categories appearing on the image would be amplified and others would be restrained. Also, the textual features are now adaptive for different input images in a non-parametric manner, other than being fixed in all existing methods~\cite{coop,adapter,zhang2021tip}. Likewise for the image, the pixel features within foreground objects, which belong to the $K$ categories, would become more notable. Meanwhile, the spatial feature map $F_s$ provides pixel-level fine-grained information for the interaction, contributing to thorough cross-modal communication. Finally, we obtain the attention-interacted global visual feature by pooling and output the classification logits as
\begin{align}
    &F_v^a = \mathrm{Pooling}(F^a_{s}),\\
    &logits = \beta_1\cdot F_v F_t^T + \beta_2\cdot F_v F^{aT}_t + \beta_3\cdot F^a_v F_t^T,
\end{align}
where $\beta_{1\sim 3}$ denote the weights for three logits: the original CLIP's logits, visual-guided logits and textual-blended logits. 
% Normally during experiments, CALIP prefers larger $\beta_3$ for textual-blended logits $F^a_v F_t^T$. 
By aggregation, CALIP achieves favorable zero-shot performance without few-shot fine-tuning or data.

\paragraph{\textbf{Analysis.}}
There are two differences between ours and the vanilla attention mechanism. The first is parametric-free: we involve no learnable parameters during the attention processing. The vanilla attention takes as input two terms and utilizes separate learnable linear layers to map them into the attention embedding space, where one as the query and the other as key and value. In contrast, our textual and visual features have already been pre-trained to be within the same space and can discard the linear layers for projection. The other difference is bidirectional. Traditional attention only updates one of the inputs, which is projected as the query, and maintains the other the same. Our design updates both of them for better interaction. As we have removed the difference for query, key and value, both input terms, visual and textual features are symmetric and play the same roles.

\subsection{CALIP-FS with Parametric Attention}
\label{calip-fs}
\paragraph{\textbf{Motivation.}}
Although the parameter-free attention enhances CLIP's zero-shot performance on a wide range of datasets, we expect to further unleash the power of cross-modal interactions under few-shot settings. Therefore, we construct CALIP-FS by inserting several learnable linear layers before and after the attention. We freeze the pre-trained encoders of CLIP and only fine-tune the inserted layers in the cross-modal attention for training efficiency.

\paragraph{\textbf{Design Details.}}
\begin{figure*}[t!]
\begin{minipage}{\linewidth}
% \vspace{0.5cm}
\centering
\begin{minipage}[t]{0.25\linewidth}
\begin{adjustbox}{width=0.96\linewidth}
\centering
\small
\begin{tabular}{c|cc}
\toprule
\multicolumn{3}{c}{\ \ \ Average over 11 2D Datasets\ \ \ } \\
% \midrule
Model & Acc. & Shot Num. \\
\midrule
CLIP & 58.53 & 0-Shot  \\
\rowcolor{blue!6}CALIP & \textbf{59.45} & 0-Shot  \\
-  & -  & - \\
\bottomrule
\end{tabular}
\end{adjustbox}
\end{minipage}
\begin{minipage}[t]{0.25\linewidth}
\begin{adjustbox}{width=0.93\linewidth}
\centering
% \small
\small
\begin{tabular}{c|cc}
\toprule
\multicolumn{3}{c}{ImageNet} \\
% \midrule
Model & Acc. & Shot Num. \\
\midrule
CLIP & 60.32 & 0-Shot  \\
\rowcolor{blue!6}CALIP & \textbf{60.57} & 0-Shot  \\
\ \ CoOp\ \  & 59.99 & \textbf{4-Shot}  \\
\bottomrule
\end{tabular}
\end{adjustbox}
\end{minipage}
\begin{minipage}[t]{0.24\linewidth}
\begin{adjustbox}{width=0.95\linewidth}
\centering
\small
\begin{tabular}{c|cc}
\toprule
\multicolumn{3}{c}{Caltech101} \\
% \midrule
Model & Acc. & Shot Num. \\
\midrule
CLIP & 83.94 & 0-Shot  \\
\rowcolor{blue!6}CALIP & \textbf{87.71} & 0-Shot  \\
CoOp & 87.53 & \textbf{1-Shot}  \\
\bottomrule
\end{tabular}
\end{adjustbox}
\end{minipage}
\begin{minipage}[t]{0.23\linewidth}
\begin{adjustbox}{width=0.99\linewidth}
\centering
\small
\begin{tabular}{c|cc}
\toprule
\multicolumn{3}{c}{SUN397} \\
% \midrule
Model & Acc. & Shot Num. \\
\midrule
CLIP & 58.53 & 0-Shot  \\
\rowcolor{blue!6}CALIP & \textbf{58.59} & 0-Shot  \\
Linear. &54.49 &\textbf{4-shot}\\
\bottomrule
\end{tabular}
\end{adjustbox}
\end{minipage}
\end{minipage}
\begin{minipage}{\linewidth}
\vspace{0.4cm}
\centering
\begin{minipage}[t]{0.25\linewidth}
\begin{adjustbox}{width=0.96\linewidth}
\centering
% \small
\small
\begin{tabular}{c|cc}
\toprule
\multicolumn{3}{c}{Food101} \\
% \midrule
Model & Acc. & Shot Num. \\
\midrule
CLIP & 77.32 & 0-Shot  \\
\rowcolor{blue!6}CALIP & \textbf{77.42} & 0-Shot  \\
CLIP-A. & 77.20 & \textbf{2-Shot}  \\
\bottomrule
\end{tabular}
\end{adjustbox}
\end{minipage}
\begin{minipage}[t]{0.25\linewidth}
\begin{adjustbox}{width=0.93\linewidth}
\centering
\small
\begin{tabular}{c|cc}
\toprule
\multicolumn{3}{c}{Flowers102} \\
% \midrule
Model & Acc. & Shot Num. \\
\midrule
CLIP & 66.10 & 0-Shot  \\
\rowcolor{blue!6}CALIP & \textbf{66.38} & 0-Shot  \\
Linear. & 58.07 & \textbf{1-Shot}  \\
\bottomrule
\end{tabular}
\end{adjustbox}
\end{minipage}
\begin{minipage}[t]{0.24\linewidth}
\begin{adjustbox}{width=0.95\linewidth}
\centering
\small
\begin{tabular}{c|cc}
\toprule
\multicolumn{3}{c}{StanfordCars} \\
% \midrule
Model & Acc. & Shot Num. \\
\midrule
CLIP & 55.71 & 0-Shot  \\
\rowcolor{blue!6}CALIP & \textbf{56.27} & 0-Shot  \\
CoOp & 55.59 &\textbf{1-Shot}  \\
\bottomrule
\end{tabular}
\end{adjustbox}
\end{minipage}
\begin{minipage}[t]{0.23\linewidth}
\begin{adjustbox}{width=0.99\linewidth}
\centering
\small
\begin{tabular}{c|cc}
\toprule
\multicolumn{3}{c}{FGVCAircraft} \\
% \midrule
Model & Acc. & Shot Num. \\
\midrule
CLIP & 17.10 & 0-Shot  \\
\rowcolor{blue!6}CALIP & \textbf{17.76} & 0-Shot  \\
CoOp& 9.64 & \textbf{1-Shot}  \\
\bottomrule
\end{tabular}
\end{adjustbox}
\end{minipage}
\end{minipage}
\begin{minipage}{\linewidth}
\vspace{0.4cm}
\centering
\begin{minipage}[t]{0.25\linewidth}
\begin{adjustbox}{width=0.96\linewidth}
\centering
\small
\begin{tabular}{c|cc}
\toprule
\multicolumn{3}{c}{OxfordPets} \\
% \midrule
Model & Acc. & Shot Num. \\
\midrule
CLIP & 85.83 & 0-Shot  \\
\rowcolor{blue!6}CALIP & \textbf{86.21} & 0-Shot  \\
\ CoOp\  & 85.32 & \textbf{8-Shot}  \\
\bottomrule
\end{tabular}
\end{adjustbox}
\end{minipage}
\begin{minipage}[t]{0.25\linewidth}
\begin{adjustbox}{width=0.93\linewidth}
\centering
\small
\begin{tabular}{c|cc}
\toprule
\multicolumn{3}{c}{DTD} \\
% \midrule
Model & Acc. & Shot Num. \\
\midrule
CLIP & 40.07 & 0-Shot  \\
\rowcolor{blue!6}CALIP & \textbf{42.39} & 0-Shot  \\
Linear. & 39.48 & \textbf{2-Shot}  \\
\bottomrule
\end{tabular}
\end{adjustbox}
\end{minipage}
\begin{minipage}[t]{0.24\linewidth}
\begin{adjustbox}{width=0.95\linewidth}
\centering
\small
\begin{tabular}{c|cc}
\toprule
\multicolumn{3}{c}{EuroSAT} \\
% \midrule
Model & Acc. & Shot Num. \\
\midrule
CLIP & 37.54 & 0-Shot  \\
\rowcolor{blue!6}CALIP & \textbf{38.90} & 0-Shot  \\
- & - & -  \\
\bottomrule
\end{tabular}
\end{adjustbox}
\end{minipage}
\begin{minipage}[t]{0.23\linewidth}
\begin{adjustbox}{width=0.99\linewidth}
\centering
\small
\begin{tabular}{c|cc}
\toprule
\multicolumn{3}{c}{UCF101} \\
% \midrule
Model & Acc. & Shot Num. \\
\midrule
CLIP & 61.33 & 0-Shot  \\
\rowcolor{blue!6}CALIP & \textbf{61.72} & 0-Shot  \\
Linear. & 53.55 & \textbf{2-Shot}  \\
\bottomrule
\end{tabular}
\end{adjustbox}
\end{minipage}
\vspace{0.1cm}
\end{minipage}
\tabcaption{\textbf{Zero-Shot Performance (\%) of CALIP on Eleven 2D Datasets.} Our zero-shot CALIP can consistently outperform CLIP and \textbf{even surpass some methods with few-shot fine-tuning.} ``Linear.'' and ``CLIP-A.'' denote Linear-probe CLIP and CLIP-Adapter, respectively.}
% \vspace{0.2cm}
\label{2dcalip}
\end{figure*}
\begin{figure*}[t!]
\begin{minipage}{\linewidth}
\vspace{0.1cm}
\centering
\begin{minipage}[t]{0.24\linewidth}
\begin{adjustbox}{width=0.95\linewidth}
\centering
\small
\begin{tabular}{c|cc}
\toprule
\multicolumn{3}{c}{Average over 3 3D Datasets} \\
% \midrule
Model & Acc. & Shot Num. \\
\midrule
PointCLIP & 21.90 & 0-Shot  \\
\rowcolor{orange!7}CALIP & \textbf{23.60} & 0-Shot  \\
\bottomrule
\end{tabular}
\end{adjustbox}
\end{minipage}
\begin{minipage}[t]{0.24\linewidth}
\begin{adjustbox}{width=0.95\linewidth}
\centering
\small
\begin{tabular}{c|cc}
\toprule
\multicolumn{3}{c}{ModelNet10} \\
% \midrule
Model & Acc. & Shot Num. \\
\midrule
PointCLIP & 30.13 & 0-Shot  \\
\rowcolor{orange!7}CALIP &\textbf{32.44} & 0-Shot  \\
\bottomrule
\end{tabular}
\end{adjustbox}
\end{minipage}
\begin{minipage}[t]{0.24\linewidth}
\begin{adjustbox}{width=0.95\linewidth}
\centering
\small
\begin{tabular}{c|cc}
\toprule
\multicolumn{3}{c}{ModelNet40} \\
% \midrule
Model & Acc. & Shot Num. \\
\midrule
PointCLIP & 20.18 & 0-Shot  \\
\rowcolor{orange!7}CALIP &\textbf{21.47} & 0-Shot  \\
\bottomrule
\end{tabular}
\end{adjustbox}
\end{minipage}
\begin{minipage}[t]{0.24\linewidth}
\begin{adjustbox}{width=\linewidth}
\centering
\small
\begin{tabular}{c|cc}
\toprule
\multicolumn{3}{c}{ScanObjectNN} \\
% \midrule
Model & Acc. & Shot Num. \\
\midrule
PointCLIP & 15.38 & 0-Shot  \\
\rowcolor{orange!7}CALIP &\textbf{16.90} & 0-Shot  \\
\bottomrule
\end{tabular}
\end{adjustbox}
\end{minipage}
\end{minipage}
\tabcaption{\textbf{Zero-Shot Performance (\%) of CALIP on Three 3D Datasets.} We extend CALIP for 3D point cloud recognition based on PointCLIP under zero-shot settings, where CALIP shows stable performance enhancement.}
\label{3dcalip}
% \vspace{0.1cm}
\end{figure*}

As shown in Figure~\ref{calip_p2},
to save the parameters, we apply a modal-shared pre-projection layers to transform the textual feature $F_t$ and spatial visual feature $F_s$ into the $C$-dimensional query, key and value,
\begin{align}
    &Q_t, K_t, V_t = \mathrm{PreProject}(F_t),\\
    &Q_s, K_s, V_s = \mathrm{PreProject}(F_s),
\end{align}
where $\mathrm{PreProject}(\cdot)$ is composed of three linear layers respectively for query, key and value and shared for two modalities. Then, we calculate two attention maps,
\begin{align}
    &A_t = \mathrm{SoftMax}(\frac{Q_t K_s^T}{\sqrt{C}})\ \in R^{K\times HW},\\
    &A_s = \mathrm{SoftMax}(\frac{Q_s K_t^T}{\sqrt{C}})\ \in R^{HW\times K},
\end{align}
where $A_t$ and $A_s$ are respectively for textual and visual features update. As the learnable projection layers are available, we could specify the attention maps to achieve modal-specific attention calculation. Afterwards, we obtain the updated features with shared post-projection layers,
\begin{align}
    &F^a_t = \mathrm{PostProject}(A_t V_s),\\
    &F^a_s = \mathrm{PostProject}(A_s V_t),
    % &F^a_v = Pooling(F^a_s)\\
    % &logits = \beta_1\cdot F_v F_t^T + \beta_2\cdot F_v F^{aT}_t + \beta_3\cdot F^a_v F_t^T,
\end{align}
where $\mathrm{PostProject}(\cdot)$ only contains one linear layer. Then, we apply pooling to process $F_t^a$ and acquire the final predicted logits by weighted summation of three terms, the same as the non-parametric version above. Equipped with such learnable projection layers, CALIP-FS significantly improves the performance over zero-shot CALIP and achieves competitive results among other state-of-the-art models by few-shot fine-tuning.

\section{Experiments}
\subsection{Zero-shot CALIP}
\label{zero-shot}

\begin{figure*}[t]
\vspace{0.4cm}
  \centering
  \includegraphics[width=0.9\textwidth]{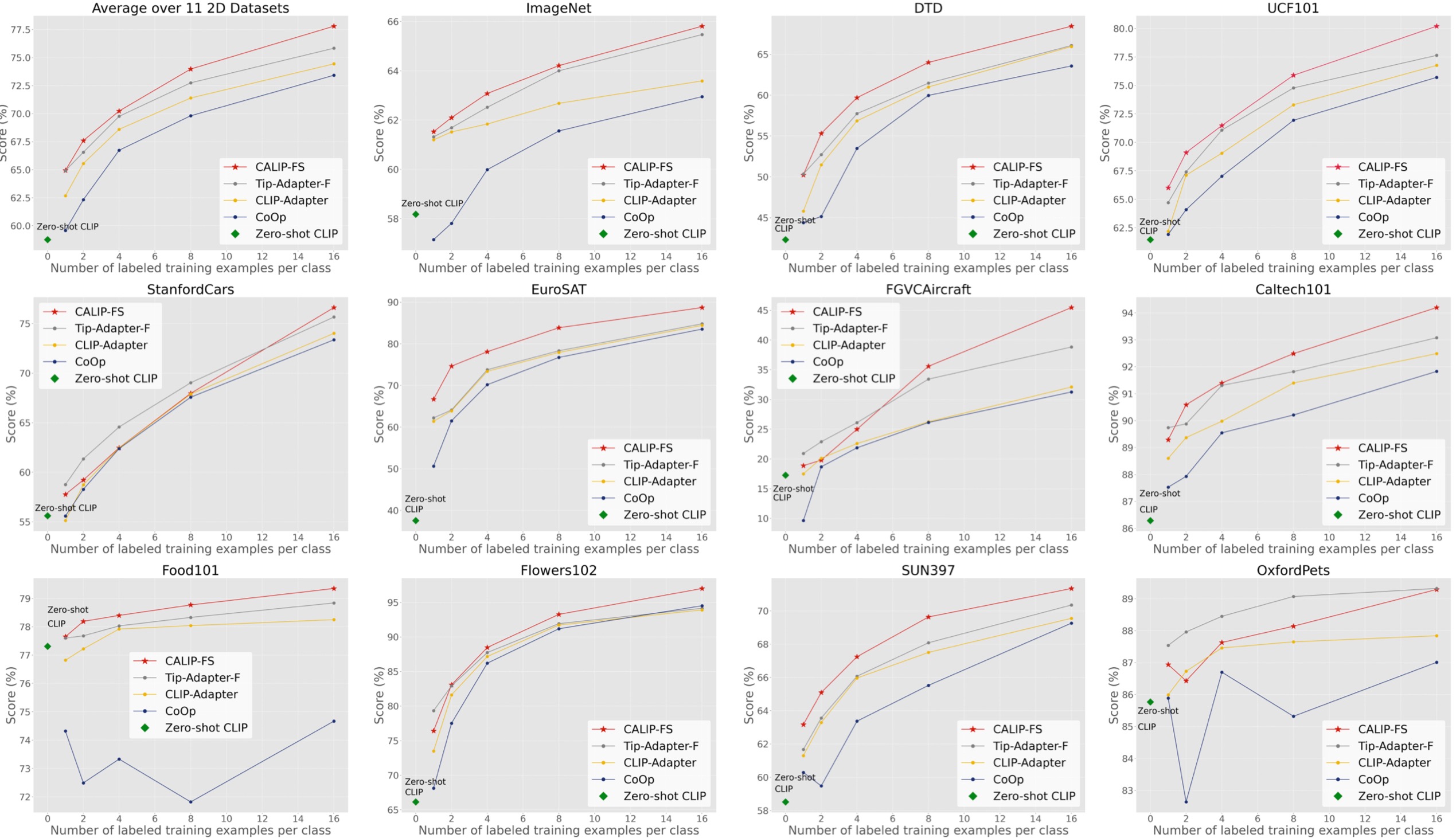}
  \caption{Few-shot Performance pf CALIP-FS on Eleven 2D Datasets. CALIP-FS shows the overall best performance over previous baselines for few-shot recognition of a wide range of visual concepts.}
  \label{2dcalip_fs}
%   \vspace{-0.2cm}
\end{figure*}

\paragraph{\textbf{Datasets}}
To fully evaluate the zero-shot enhancement of CALIP, we experiment on a wide range of benchmarks including 11 image 2D datasets and 3 point cloud 3D datasets. 2D datasets contain a variety of visual concepts, e.g., real-world scenarios, satellite-captured landscapes and detailed textures, which are ImageNet~\cite{imagenet}, Caltech101 ~\cite{feifei04}, OxfordPets~\cite{parkhi}, StanfordCars~\cite{krause}, Flowers102~\cite{nizi}, Food101~\cite{bossard}, FGVCAircraft~\cite{maji}, SUN397~\cite{xiao10}, DTD~\cite{cimpoi}, EuroSAT~\cite{helber} and UCF101~\cite{soomro}. The 3D datasets include both synthetic and sensor-scanned point clouds: ModelNet10~\cite{modelnet}, ModelNet40~\cite{modelnet} and ScanObjectNN~\cite{scan}. As CALIP requires no downstream data for training, we utilize no training sets of the datasets and directly evaluate on their full test sets.

\paragraph{\textbf{Settings}}
We adopt ResNet-50~\cite{resent} as the visual encoder and a 12-layer transformer as the textual encoder. Following CLIP's~\cite{clip} pre-processing, we resize all test images into 224$\times$224 resolutions and $H, W, C$ of visual spatial feature $F_s$ denote 7, 7, 1024. We set $\alpha_t$ and $\alpha_s$ for modulating textual and visual attention magnitude both as 2. For the pooling operation of $F_s^a$, we select the combination of maximum and average poolings for better features integration. We adopt varying $\beta_1, \beta_2, \beta_3$ for different datasets to adapt their specific domains. As for textual templates, we refer to CLIP adopting handcrafted ones. Regarding 3D point cloud recognition, CALIP follows PointCLIP~\cite{zhang2021pointclip} to project point clouds onto 6-view depth maps with the distance 1.2 and aggregate view-wise zero-shot predictions as the final output.

\paragraph{\textbf{Analysis}}
As shown in Table~\ref{2dcalip}, we compare zero-shot CALIP with CLIP and some few-shot models for 2D image classification. Our CALIP with parameter-free attention consistently outperforms CLIP on all downstream benchmarks by +0.92$\%$ average accuracy. We largely surpass CLIP by +3.77$\%$ on Caltech101 and +1.36$\%$ on EuroSAT. CALIP without training even beats existing learnable methods under few-shot fine-tuning, e.g., \textbf{surpassing 1-shot Linear-probe CLIP by +8.89$\%$ on Flowers102, and 8-shot CoOp by +0.89$\%$ on OxfordPets.} As for 3D point cloud classification in Table~\ref{3dcalip}, CALIP also enhances PointCLIP on 3 datasets by +1.70$\%$ average accuracy without parameters.

\begin{table}[t]
\vspace*{0.2cm}
\centering
\small
\begin{tabular}{lcccccc}
% \vspace*{-0.1cm}
\toprule
\multirow{3}{*}{Datasets} & \textbf{Source} &\multicolumn{4}{c}{\textbf{Target}} \\
\cmidrule(lr){2-2} \cmidrule(lr){3-6} 
& ImageNet  & -V2 &-A &-R & -Sketch \\
\midrule
CLIP  & 60.32  & 53.27 &23.61 &60.42 & 35.44\\
CALIP &\textbf{60.57}    &\textbf{53.70}   &\textbf{23.96} &\textbf{60.81} &\textbf{35.61}  \\
\cmidrule(lr){1-6}
Linear-probe  & 56.13  & 45.61 &12.71 &34.86 & 19.13\\
CoOp & 62.95  & 54.58 &23.06 &54.96 & 31.04  \\
CALIP-FS & \textbf{65.81}  & \textbf{55.98} & \textbf{23.42} & \textbf{56.74} & \textbf{35.37} \\
\bottomrule
\end{tabular}
% \vspace*{6pt}
\caption{{Performance (\%) on Distribution Shift.}}
% \vspace*{-0.2cm}
\label{dg}
\end{table}

% V2:53.70
% Ske: 35.61
% r:60.42 -> 60.81
% a:23.61 -> 23.96

\subsection{Few-shot CALIP-FS}
\paragraph{\textbf{Datasets}}
We evaluate CALIP-FS for few-shot classification on 11 2D datasets mentioned above and compare ours with the state-of-the-art methods: zero-shot CLIP~\cite{clip}, CoOp~\cite{coop}, CLIP-Adapter~\cite{adapter} and Tip-Adapter-F~\cite{zhang2021tip}. We follow the widely-adopted few-shot protocols, which randomly sample 1, 2, 4, 8 and 16 shots of each category for training and test models on the full test set.

% \paragraph{\textbf{Settings}}
% For few-shot experiments, unified training settings are adopted on all datasets. We train CALIP-FS for 200 epochs of batch size 32 and utilize SGD optimizer with the learning rate $2e^{-3}$. We also apply ResNet-50\cite{resent} and the 12-layer transformer~\cite{vaswani} as the visual and textual encoders as default. The parametric attention module is conducted only one time and its number of heads is set as 4. Instead of using learnable continuous textual templates in CoOp, we inherit hand-crafted ones in zero-shot CALIP. In regard to data augmentation, we choose random crop and resize augmented images into 224$\times$224 resolutions.
% For other methods to be compared, we respectively select the best-performing variants presented in their papers. Therein, we report CoOp with 16-length learnable texts, which shares tokens over categories and places the [CLASS] token at the end of textual inputs. For zero-shot CLIP, CLIP-Adapter and Tip-Adapter, we adopt the same handcrafted templates as CALIP-FS. For Tip-Adapter, we actually report Tip-Adapter-F, which fine-tunes the constructed cache model for better few-shot performance. Note that the linear probe of CLIP performs much worse than existing methods as reported in~\cite{coop}, so we omit their results in the comparison.

% \vspace{-0.1cm}
\paragraph{\textbf{Analysis}}
The main results are presented in Figure~\ref{2dcalip_fs}. The average accuracy over 11 datasets on the top-left corner indicates CALIP-FS's superior few-shot performance over all other baselines.
Based on zero-shot CLIP, CALIP-FS achieves significant performance improvements, especially on DTD and EuroSAT, ranging from +20$\%$ to +50$\%$. Compared to other few-shot methods, we only lag behind Tip-Adapter-F on OxfordPets, and largely outperform others on DTD, EuroSAT and SUN397. More importantly, rather than Tip-Adapter-F's complicated two-step fine-tuning by storing all training samples, CALIP-FS is more efficient and simple with the one-step training.

% Compared with \textbf{CoOp}\cite{zhou21}, our VT-CLIP outperforms CoOp under the number of shots ranging from 1 to 16. Although CoOp achieve considerable improvement over Zero-shot CLIP, the performance are still worse than our VT-CLIP. Note that CoOp achieve the result from a perspective of prompt learning and demonstrate that the prompt base methods are competitive approach for enhancing vision-language model but its potential is not more promising than finetuning an additional module over frozen pre-trained model.

% Compared with \textbf{CLIP-Adapter}\cite{clip}, VT-CLIP performs better than CLIP-Adapter in average accuracy of all datasets and under 16-shot setting, VT-CLIP achieves higher accuracy over CLIP-Adapter except for a slight degradation in DTD and EuroSAT. Although CLIP-Adapter is a strong competitor that trains a adapter module over pre-trained model which is similar to our methods in perspective of “pretrain-finetuning” paradigm, our VT-CLIP achieves higher performance and the key difference is that VT-CLIP leverage the contextual visual feature to guided the text feature, which enable the text feature to become more semantically correlated to the downstream tasks. The experiment results demonstrate that our VT-CLIP with intersection between visual and text branch of CLIP is a more promising perspective of enhancing the vision-language model.
\begin{figure*}[!t]
% \vspace{1cm}
  \centering
    \includegraphics[width=0.9\textwidth]{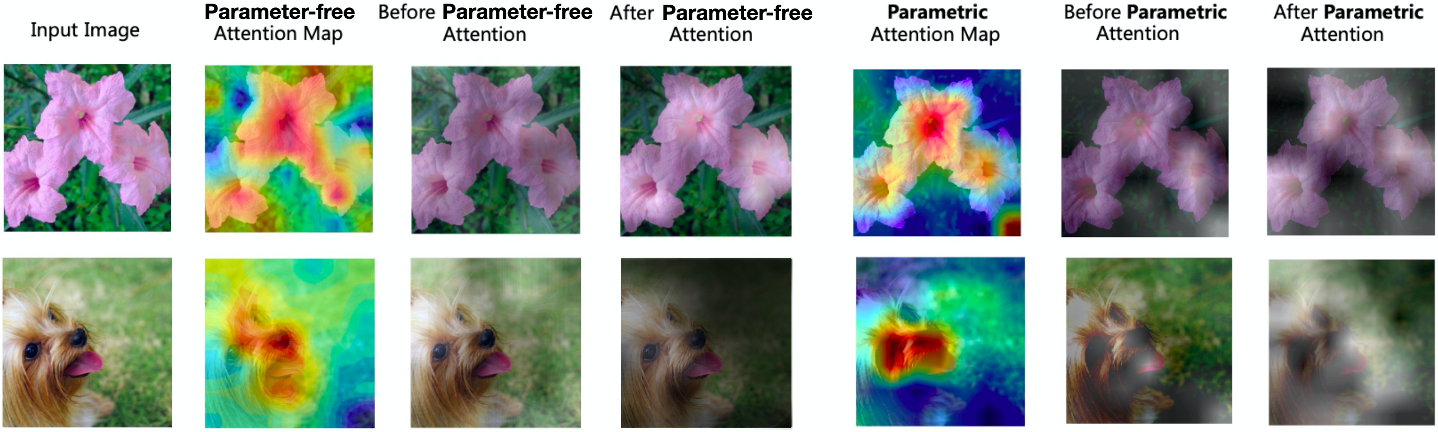}  \figcaption{Visualization of Attention Maps and Spatial Visual Features in CALIP and CALIP-FS.}
    \label{calip_p3}
    % \vspace{-0.3cm}
\end{figure*}

\subsection{Out-of-distribution Performance}
Robustness to distribution shift is a common benchmark to evaluate the generalization ability of deep-learning models. 
We evaluate the out-of-distribution performance of CALIP and CALIP-FS by training on ImageNet and testing on ImageNetV2~\cite{recht2019imagenet}, ImageNet-Sketch~\cite{wang2019learning}, ImageNet-A~\cite{hendrycks2021natural} and ImageNet-R~\cite{hendrycks2021many}. These test datasets contain compatible categories with ImageNet but within different visual domains. In Table~\ref{dg}, we compare ours with the published results of zero-shot CLIP, Linear-probe CLIP and CoOp. As shown, CALIP acquires better generalization ability than CLIP without training. By 16-shot fine-tuning, CALIP-FS also surpasses CoOp on four out-of-distribution datasets.

\begin{table}[t]
% \vspace{1cm}
\centering
\begin{adjustbox}{width=\linewidth}
	\begin{tabular}{lc c cccc}
	\toprule
		\multicolumn{6}{c}{Combination of Logits} \\
		\cmidrule(lr){1-6}
	$F_v F_t^T$ &$F_v F^{aT}_t$ &$F^a_v F_t^T$ &$F_v^a F^{aT}_t$ &CALIP &CALIP-FS\\
		 \cmidrule(lr){1-4} \cmidrule(lr){5-6} 
		 \checkmark&-&-&-&83.94\%&83.94\%\\
		 \checkmark &\checkmark &-&-&84.02\%&94.42\%\\
		 \checkmark &-&\checkmark &-&85.10\%&93.39\%\\
		 \checkmark &-&-&\checkmark &81.96\%&94.60\%\\
		 \checkmark &\checkmark &\checkmark &-&\textbf{ 85.66\%}&\textbf{ 94.75\%}\\
		 \checkmark&\checkmark &\checkmark &\checkmark&85.34\% &94.66\%\\
	\bottomrule
	\end{tabular}
\end{adjustbox}
\vspace*{1pt}
\caption{Ablation Study of Logits Combination.}
\label{ablation1}
% \vspace*{0.1cm}
\end{table}

\subsection{Ablation Study}
To further demonstrate the theory of our approach, we conduct ablation studies on Caltech101 dataset with zero-shot CALIP and 16-shot CALIP-FS. We report our results on the official validation set for tuning hyperparameters and network structures.
% For other compared models, Linear-probe CLIP and CoOp, we also train them by 16-shot datasets.

% \begin{table}[h]
% % \vspace{1cm}
% \centering
% \begin{adjustbox}{width=\linewidth}
% 	\begin{tabular}{lc c cccc}
% 	\toprule
% 		\multicolumn{6}{c}{Combination of Logits} \\
% 		\cmidrule(lr){1-6}
% 	$F_v F_t^T$ &$F_v F^{aT}_t$ &$F^a_v F_t^T$ &$F_v^a F^{aT}_t$ &CALIP &CALIP-FS\\
% 		 \cmidrule(lr){1-4} \cmidrule(lr){5-6} 
% 		 \checkmark&-&-&-&83.94\%&83.94\%\\
% 		 \checkmark &\checkmark &-&-&84.91\%&92.33\%\\
% 		 \checkmark &-&\checkmark &-&87.10\%&93.55\%\\
% 		 \checkmark &-&-&\checkmark &83.96\%&93.48\%\\
% 		 \checkmark &\checkmark &\checkmark &-&\textbf{87.71\%}&\textbf{93.63\%}\\
% 		 \checkmark&\checkmark &\checkmark &\checkmark&87.34\% &93.43\%\\
% 	\bottomrule
% 	\end{tabular}
% \end{adjustbox}
% \vspace*{3pt}
% \caption{\textbf{Ablation Study of Logits Combination.}}
% \label{ablation1}
% \vspace*{-0.3cm}
% \end{table}

\paragraph{\textbf{Cross-modal Attention}}
The attention aggregates three terms of logits for the final output: $F_v F_t^T$, $F_v F^{aT}_t$ and $F^a_v F_t^T$, where the first term is the CLIP's original prediction and the other two respectively contains the attention-interacted $F^a_t$ and $F^a_v$. There actually exists the fourth term: $F_v^a F^{aT}_t$, that is, the logits predicted by the updated features of both modalities. In Table~\ref{ablation1}, we explore their best combination form and observe that, for both CALIP and CALIP-FS, the fourth term $F_v^a F^{aT}_t$ would adversely influence the predicted logits, since its too much cross-modal interaction might harm the already well-aligned knowledge from pre-trained CLIP. In contrast, the combination of logits that only interact one modality via the attention performs better. It not only preserves the effective pre-trained CLIP's knowledge, but also fuses newly-interacted cross-modal knowledge.

% \paragraph{\textbf{Visual Encoders}}
% We vary the visual encoders in CALIP and CALIP-FS with various alternatives to evaluate their performance. In Table~\ref{encoder}, we list the results of CLIP and CoOp for comparison, which is the only available data we could access. As shown, with different visual networks, CALIP and CALIP-FS still achieves better performance than CLIP and CoOp, respectively.

\paragraph{\textbf{Pre/Post-Projection Layers}}
We explore where to insert learnable linear layers in CALIP's parameter-free attention to construct CALIP-FS. As shown in Table~\ref{projection}, equipping both pre/post-projection layers for two modalities achieves the best performance. This design decouples the embedding space of attention calculation from the previous one by the former projecting-in and the latter projecting-out layers, which produces better attention map for interactions.

\begin{table}[t!]
% \vspace{0.4cm}
\small
\centering
% \begin{adjustbox}{width=\linewidth}
	\begin{tabular}{ccccc}
	\toprule
% 		\multicolumn{5}{c}{Projection Design for CALIP-FS} \\
		\multicolumn{2}{c}{Visual
		Projection}&\multicolumn{2}{c}{Textual Projection}&\makecell*[c]{\multirow{2}*{Accuracy}}\vspace{2pt}\\
				\cmidrule(lr){1-4}
	Pre-Proj. &Post-Proj. &Pre-Proj. &Post-Proj. \vspace{2pt}\\
		 \cmidrule(lr){1-2}\cmidrule(lr){3-4} \cmidrule(lr){5-5} 
		 - &- &-&- &87.71\%\\
		 \checkmark &-&\checkmark&-   &89.75\%\\
		 \checkmark &- &\checkmark &\checkmark &90.36\%\\
		 \checkmark &\checkmark &\checkmark &-&93.94\%\\
		 \checkmark&\checkmark&\checkmark&\checkmark&\textbf{ 94.75\%} \\
	\bottomrule
	\end{tabular}
% \end{adjustbox}
\vspace*{2pt}
\caption{Ablation Study of Pre/Post-Projection Designs.}
\label{projection}
% \vspace*{0.1cm}
\end{table}

\section{Visualization}
In Figure~\ref{calip_p3}, we visualize attention maps, spatial visual features before and after the CALIP's parameter-free attention and CALIP-FS's parametric attention, respectively.
As shown, for both variants, the attention maps concentrate well around the object pixels, and the visual features become more distinctive guided by category texts as expected. Also, after few-shot fine-tuning, the distributions of attention maps and visual features all get more intensive, which indicates the improvements resulted from learnable parameters.

% We use the heat map to visualize the attention map of the cross attention module after training CALIP-FS on two images of \textit{Airliner} from ImageNet to show the learned characteristic of CALIP-FS. The result is displayed in \cref{fig:visualization}. For attention map, the more reddish a region in the heat map is, the more attention is paid to that region. We can observe that the attention of CALIP-FS is focus on the engines of the airline which is a vital characteristic for airline. We also display the predicted logits of CALIP-FS and zero-shot CLIP for the same example. We can observe that the logits from the CALIP-FS is more concentrated in the ground truth category than Zero-shot CLIP while the score for other categories are less than Zero-shot CLIP. From the comparison between Zero-shot CLIP and our CALIP-FS, we claim that our CALIP-FS is more effective in recognizing the ground true category among
% the similar classes. In summary, the visualization results prove that visual guided text feature pay more attention to the informative region of the image and enhance the vision-language model under few-shot setting.

\section{Conclusion}
We propose CALIP, the first work to conduct zero-shot enhancement over CLIP via a parameter-free attention module. CALIP interacts visual and textual features without any parameters or training and achieves favorable performance over a wide range of 2D and 3D benchmarks. Then, we introduce the parametric version CALIP-FS to further boost its classification accuracy under few-shot fine-tuning and acquire competitive results among existing state-of-the-art methods. We hope our work could inspire future researches for zero-shot enhancement of pre-trained large-scale multi-modal models. Concerning limitations, we will further extend our parameter-free methods for CLIP-based object detection and semantic segmentation.

\section*{Acknowledgements}
This work is supported by NSFC (No. 61832001 and U22B2037).

\section{Few-shot Training Settings}

\begin{table}[t!]
\small
\centering
% \vspace*{-0.4cm}
\begin{adjustbox}{width=\linewidth}
 \centering
 \begin{tabular}{lccccccc}
 \toprule
 \multicolumn{7}{c}{$\beta_2, \beta_3$ for CALIP-FS} \\
 \toprule
 \specialrule{0em}{1pt}{1pt}
 \ \ \ $\beta{_2}$ &0.08 &0.1 &\textbf{0.12} &0.14 &0.16 &0.18\\
 \cmidrule(lr){1-1} \cmidrule(lr){2-7}
 \specialrule{0em}{1pt}{1pt}
 \ \ Test &92.33\% &92.86\% &\textbf{94.20\%} &93.59\% &93.50\% &93.54\%\\ 
 \specialrule{0em}{1pt}{1pt}
 \ \ Val &92.50\% &93.25\% &\textbf{93.50\%} &93.25\% &93.25\% &93.25\%\\ 
\specialrule{0em}{1pt}{1pt}
 \toprule
 \specialrule{0em}{1pt}{1pt}
 \ \ \ $\beta{_3}$ &0.08 &0.1 &\textbf{0.12} &0.14 &0.16 &0.18\\
 \cmidrule(lr){1-1} \cmidrule(lr){2-7}
 \specialrule{0em}{1pt}{1pt}
 \ \ Test &92.98\% &93.18\% &\textbf{94.20\%} &93.39\% &93.35\% &93.14\%\\ 
 \specialrule{0em}{1pt}{1pt}
 \ \ Val &93.25\% &93.25\% &\textbf{93.50\%} &93.50\% &93.25\% &93.50\%\\ 
\specialrule{0em}{1pt}{1pt}
 \bottomrule
 \end{tabular}
\end{adjustbox}
% \vspace*{6pt}
\caption{\textbf{Ablation Study of Logits Weights $\beta_2, \beta_3$ for 16-shot CALIP-FS.}}
% \vspace*{-5pt}
\label{betacalipfs_both}
\end{table}

\begin{table}[t!]
\vspace{2pt}
\small
\begin{adjustbox}{width=\linewidth}
	\centering
	\begin{tabular}{lccccccc}
	\toprule
		\multicolumn{7}{c}{$\beta_2, \beta_3$ for CALIP} \\
		\midrule
		\specialrule{0em}{1pt}{1pt}
		 \ \ \ $\beta{_2}$  &3.50 &4.00 &4.50 &\textbf{5.00} &5.50 &6.00\\
        \cmidrule(lr){1-1} \cmidrule(lr){2-7}
        \specialrule{0em}{1pt}{1pt}
		 \ \ Test &87.14\%  &87.10\%  &86.94\%  &\textbf{87.51\%}  &86.57\%  &86.41\%  \\ 
		 \specialrule{0em}{1pt}{1pt}
		 \ \ Val  &82.75\%  &82.50\%  &82.75\%  &\textbf{83.25\%}  &82.50\%  &83.00\%  \\ 
		 \specialrule{0em}{1pt}{1pt}
		 \toprule
		 \specialrule{0em}{1pt}{1pt}
		 \ \ \ $\beta{_3}$  &0.14 &0.16 &\textbf{0.18} &0.20 &0.22 &0.24\\
        \cmidrule(lr){1-1} \cmidrule(lr){2-7}
        \specialrule{0em}{1pt}{1pt}
		 \ \ Test &86.82\%  &87.06\%  &\textbf{87.51\%}  &86.98\%  &86.45\%  &86.29\%    \\ 
		 \specialrule{0em}{1pt}{1pt}
		 \ \ Val &83.00\%  &83.25\%  &\textbf{83.25\%}  &82.75\%  &82.00\%  &81.75\%    \\ 
		 \specialrule{0em}{1pt}{1pt}
	\bottomrule
	\end{tabular}
\end{adjustbox}
% \vspace*{6pt}
\caption{\textbf{Ablation Study of Logits Weights $\beta_2, \beta_3$ for Zero-shot CALIP.}}
% \vspace*{-5pt}
\label{betacalip_both}
\end{table}

We conduct few-shot experiments concerning CoOp~\cite{coop}, CLIP-Adapter~\cite{adapter}, Tip-Adapter-F~\cite{zhang2021tip} and our CALIP-FS on 11 2D datasets under 1, 2, 4, 8 and 16 shots' fine-tuning. Note that Linear-probe CLIP performs much worse than other existing methods as reported in~\cite{coop}, so we omit their results for comparison in the main paper. Specifically, we train CALIP-FS for 200 epochs of batch size 32 and utilize SGD optimizer with the learning rate $2e^{-3}$. We apply ResNet-50\cite{resent} and the 12-layer transformer~\cite{vaswani} for the visual and textual encoders as default. Instead of using learnable continuous textual templates in CoOp, we inherit handcrafted templates the same as CLIP, zero-shot CALIP, CLIP-Adapter and Tip-Adapter. In regard to data augmentation, we follow existing works to use random crop with 224 $\times$ 224 image resolutions. 
For other methods to be compared, we respectively select their best-performing variants presented in the papers. Therein, we report CoOp with 16-length learnable texts, which shares the same token sequence over all categories and places the [CLASS] token at the end. For CLIP-Adapter, we compare the model only with the visual adapter over the visual encoder, which achieves better performance than adapters of both modalities.
For Tip-Adapter, we report the results of Tip-Adapter-F, which fine-tunes the constructed cache model and largely surpasses the training-free Tip-Adapter.

\begin{table}[h!]
\small
\centering
% \vspace*{0.4cm}
\begin{adjustbox}{width=\linewidth}
	\centering
	\begin{tabular}{lccccccc}
	\toprule
		\multicolumn{7}{c}{$\alpha_t, \alpha_s$ for CALIP} \\
		\midrule
		\specialrule{0em}{1pt}{1pt}
		\ \ \ $\alpha_{t}$  &0.25 &0.50 &1.00 &\textbf{2.00} &4.00 &8.00\\
        \cmidrule(lr){1-1} \cmidrule(lr){2-7}
        \specialrule{0em}{1pt}{1pt}
		\ \ Test &87.46\% &87.46\%  &87.34\%  &\textbf{87.51\%}  &87.26\%  &87.38\%  \\ 
		\specialrule{0em}{1pt}{1pt}
		\ \ Val &82.25\% &83.00\%  &82.75\%  &\textbf{83.25\%}  &82.25\%  &81.75\%  \\ 
		 \specialrule{0em}{1pt}{1pt}
		 \toprule
		 \specialrule{0em}{1pt}{1pt}
		 \ \ \ $\alpha_{s}$  &1.50 &\textbf{2.00} &2.50 &3.00 &3.50 &4.00\\
        \cmidrule(lr){1-1} \cmidrule(lr){2-7}
        \specialrule{0em}{1pt}{1pt}
		\ \ Test &87.46\%  &\textbf{87.51\%}  &87.06\%  &86.86\%  &86.73\%  &86.45\%    \\
		\specialrule{0em}{1pt}{1pt}
		\ \ Val &83.00\%  &\textbf{83.25\%}  &82.50\%  &82.75\%  &81.25\%  &82.00\%    \\
		 \specialrule{0em}{1pt}{1pt}
	\bottomrule
	\end{tabular}
\end{adjustbox}
% \vspace*{3pt}
\caption{\textbf{Ablation Study of Attention Magnitudes $\alpha_t$ and $\alpha_s$ for Zero-shot CALIP.}}
% \vspace*{8pt}
\label{modulate}
\end{table}

\begin{figure*}[t!]
  \centering
    \includegraphics[width=\textwidth]{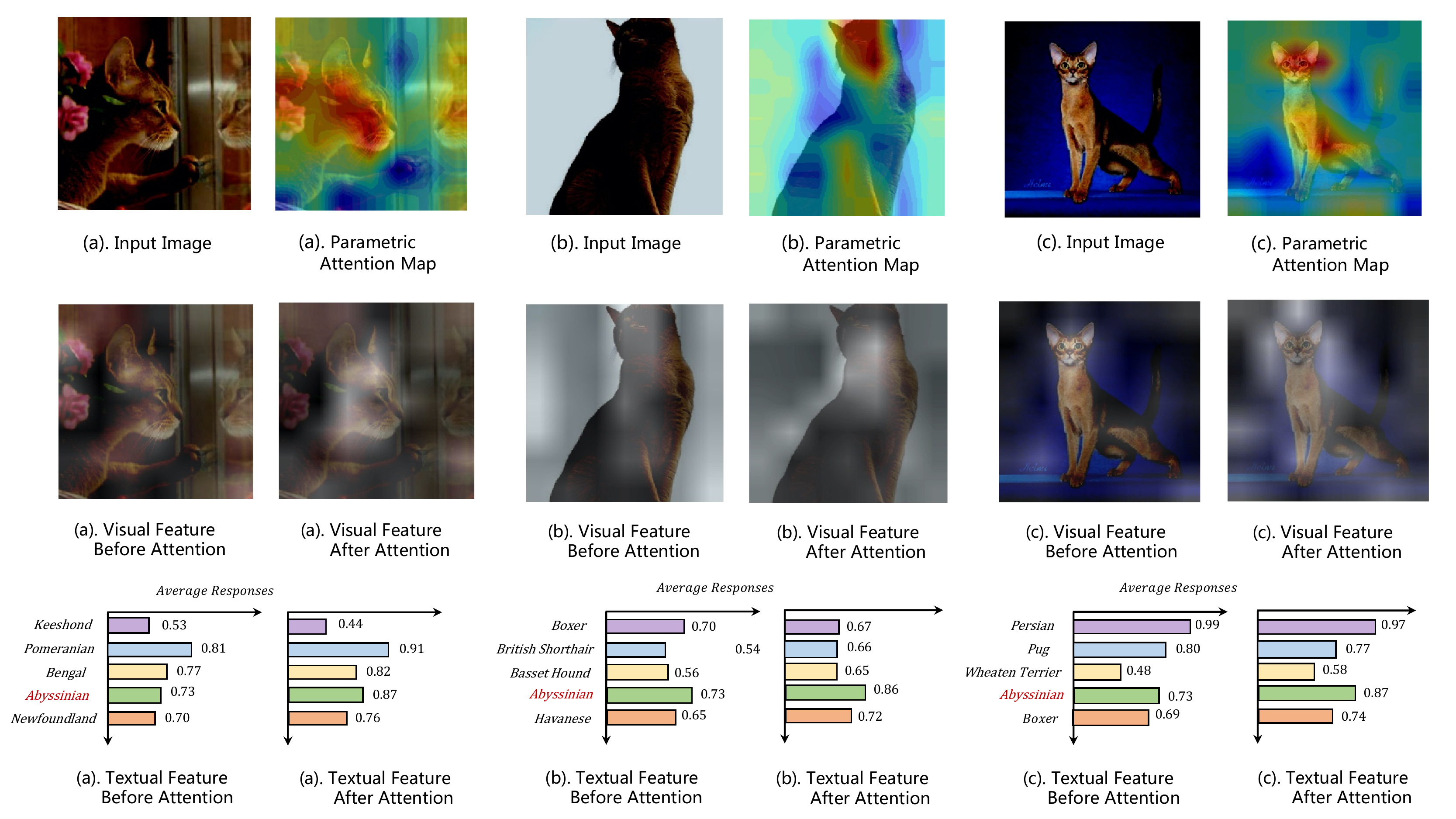}
   \caption{\textbf{Visualization of Attention Maps, Spatial Visual Features and Textual Features for CALIP-FS with Parametric Attention.} Both visual and textual features are enhanced during our cross-modal interactions.}
    \label{calip_sup}
\end{figure*}
\section{Additional Ablation Study}
To further explain the theory of our method, we conduct additional ablation study for both zero-shot CALIP and 16-shot CALIP-FS. We tune all hyperparameters on the official validation set and also report the corresponding test-set results for reference.

% We firstly analyze the effect of five hyperparameters in our model to the performance, i.e., $\beta_1, \beta_2, \beta_3, \alpha_t, \alpha_s$. Then, we experiment with different visual encoders to test our robustness of network architectures.
\begin{table}[t!]
% \vspace{0.2cm}
\small
	\centering
	\begin{adjustbox}{width=\linewidth}
	\begin{tabular}{lccccccc}
	\toprule
		\multicolumn{7}{c}{\ \ \ \ \ \ \ \ \ \ \ \ \ \ Best $\beta_1, \beta_2, \beta_3$ for All Datasets\ \ \ \ \ \ \ \ \ \ \ \ \ \ } \\
		\midrule
		\multirow{2}{*}{\ \ Datasets}
		&\multicolumn{3}{c}{CALIP}&\multicolumn{3}{c}{CALIP-FS} \\
		\cmidrule(lr){2-7}
		&\ $\beta_1$&$\beta_2$&$\beta_3$&$\ \beta_1$&$\beta_2$&$\beta_3$\\
        \cmidrule(lr){1-1} \cmidrule(lr){2-4}\cmidrule(lr){5-7}
        \specialrule{0em}{1pt}{1pt}
		 \ \ ImageNet &1.00 &1.12 &0.02 &1.00 &0.12 &0.10\\ 
        %\cmidrule(lr){1-7}
        \specialrule{0em}{1pt}{1pt}
		 \ \ Caltech101  &1.00 &5.00 &0.18 &1.00 &0.12&0.12\\ 
        %\cmidrule(lr){1-7}
        \specialrule{0em}{1pt}{1pt}
		 \ \ SUN397 &1.00 &0.43 &0.01 &1.00 &0.12&0.12\\ 
        %\cmidrule(lr){1-7}
        \specialrule{0em}{1pt}{1pt}
		 \ \ Food101 &1.00 &0.60 &0.02 &1.00 &0.08&0.10\\ 
        %\cmidrule(lr){1-7}
        \specialrule{0em}{1pt}{1pt}
		 \ \ Flowers102 &1.00 &0.50 &0.01 &1.00 &0.70&0.70\\ 
        %\cmidrule(lr){1-7}
        \specialrule{0em}{1pt}{1pt}
		 \ \ StanfordCars &1.00 &2.80 &0.01 &1.00 &0.30&0.40\\ 
        %\cmidrule(lr){1-7}
        \specialrule{0em}{1pt}{1pt}
		 \ \ FGVCAircraft &1.00 &1.30 &0.01 &1.00 &0.60&1.00\\ 
        %\cmidrule(lr){1-7}
        \specialrule{0em}{1pt}{1pt}
		 \ \ OxfordPets &1.00 &0.61 &0.01 &1.00 &0.08&0.08\\ 
        %\cmidrule(lr){1-7}
        \specialrule{0em}{1pt}{1pt}
		 \ \ DTD &1.00 &1.40&0.01 &1.00 &0.30&0.20\\ 
        % \cmidrule(lr){1-7}
        \specialrule{0em}{1pt}{1pt}
		 \ \ EuroSAT  &1.00 &6.08 &0.06 &1.00 &0.40&0.40\\ 
        \specialrule{0em}{1pt}{1pt}
		 \ \ UCF101 &1.00 &1.28 &0.01 &1.00 &0.60&0.60\\ 
	\bottomrule
	\end{tabular}
\end{adjustbox}
\vspace*{3pt}
\caption{\textbf{Best Hyperparameter Settings for CALIP and CALIP-FS on All Datasets.}}
% \vspace*{-10pt}
\label{hyper}
\end{table}
\vspace*{-2pt}
\subsection{Logits Weights $\beta_1, \beta_2, \beta_3$}
For simplicity, we fix $\beta_1$ of CLIP's original $F_v F_t^T$ to be 1 in all experiments, serving as the relative base for other two logits. Here, we explore how $\beta_2, \beta_3$ for the other two logits $F_v F^{aT}_t, F^a_v F_t^T$ affect the final prediction accuracy on Caltech101~\cite{feifei04} dataset. For zero-shot CALIP in Table~\ref{betacalip_both}, blending more visual-guided logits $F_v F^{aT}_t$ achieves the best performance by the weight 5, where the textual features explore the informative visual features from the image via cross-modal interactions. This well demonstrates the significance of the parameter-free attention, since our 5-weight $F_v F^{aT}_t$ contributes more than the 1-weight CLIP's $F_v F_t^T$. In contrast, for CALIP-FS in Table~\ref{betacalipfs_both}, the pre-trained logits $F_v F_t^T$ plays the most important role among others.

\subsection{Sensitivity Test}
For $\beta_2, \beta_3$ that needed to be tuned, we evaluate their sensitivity on test sets for all datasets in Table~\ref{sensitive}. As reported, the variation of $\beta_2, \beta_3$ within a certain scope do not have too much influence to the final classification scores.

\subsection{Parameter-free Magnitudes $\alpha_t, \alpha_s$}
For zero-shot CALIP, we fix the magnitude modulators $\alpha_t, \alpha_s$ in the parameter-free attention module as 2 for all datasets. In Table~\ref{modulate}, we also show their influence on Caltech101~\cite{feifei04} dataset. As presented, setting both modulators as 2 for textual and visual attention maps reaches the highest zero-shot accuracy. This indicates the comparative roles of the two modalities for the bidirectional features updatate.

% \begin{table}[b!]
% \centering
% \small
% \begin{adjustbox}{width=\linewidth}
% 	\centering
% 	\begin{tabular}{lccccccc}
% 	\toprule
% 		\multicolumn{7}{c}{$\beta_1, \beta_2, \beta_3$ for CALIP} \\
% 		\midrule
% 		\ \ \ $\beta{_1}$  &0.25 &0.50 &0.75 &\textbf{1.00} &1.25 &1.50\\
%         \cmidrule(lr){1-1} \cmidrule(lr){2-7}
%         \specialrule{0em}{1pt}{1pt}
% 		 \rowcolor{gray!10}\ \ Acc. &75.38\%  &85.15\%  &86.41\%  &\textbf{87.51\%}  &87.18\%  &87.02\%  \\ 
% 		 \specialrule{0em}{1pt}{1pt}
% 		 \toprule
% 		 \ \ \ $\beta{_2}$  &3.50 &4.00 &4.50 &\textbf{5.00} &5.50 &6.00\\
%         \cmidrule(lr){1-1} \cmidrule(lr){2-7}
%         \specialrule{0em}{1pt}{1pt}
% 		 \rowcolor{gray!10}\ \ Acc. &87.14\%  &87.10\%  &86.94\%  &\textbf{87.51\%}  &86.57\%  &86.41\%  \\ 
% 		 \specialrule{0em}{1pt}{1pt}
% 		 \toprule
% 		 \ \ \ $\beta{_3}$  &0.14 &0.16 &\textbf{0.18} &0.20 &0.22 &0.24\\
%         \cmidrule(lr){1-1} \cmidrule(lr){2-7}
%         \specialrule{0em}{1pt}{1pt}
% 		 \rowcolor{gray!10}\ \ Acc. &86.82\%  &87.06\%  &\textbf{87.51\%}  &86.98\%  &86.45\%  &86.29\%    \\ 
% 		 \specialrule{0em}{1pt}{1pt}
% 	\bottomrule
% 	\end{tabular}
% \end{adjustbox}
% \vspace*{6pt}
% \caption{Ablation Study of Logits Weights $\beta_1, \beta_2, \beta_3$ in Zero-Shot CALIP with Non-Parametric Attention.}
% % \vspace*{-5pt}
% \label{betacalip}
% \end{table}

\begin{table}[t!]
\small
\centering
\begin{adjustbox}{width=\linewidth}
\begin{tabular}{lllll}
\toprule
Dataset        & ${\beta_2}$ Range     & Acc. Range               & ${\beta_3}$ Range     & Acc. Range               \\ 
\cmidrule(lr){1-1} \cmidrule(lr){2-2} \cmidrule(lr){3-3} \cmidrule(lr){4-4} \cmidrule(lr){5-5}
ImageNet       & 0.08$\sim$0.18 & 63.99$\sim$64.53 & 0.04$\sim$0.24 & 64.11$\sim$64.53 \\ 
DTD            & 0.20$\sim$0.30   & 67.66$\sim$68.19 & 0.20$\sim$0.30   & 67.72$\sim$68.19 \\ 
EuroSAT        & 0.40$\sim$0.50   & 87.86$\sim$88.05 & 0.30$\sim$0.40   & 88.05$\sim$88.06 \\ 
Food101        & 0.10$\sim$0.20   & 79.09$\sim$79.32 & 0.04$\sim$0.14 & 78.67$\sim$79.32 \\ 
Flowers102 & 0.60$\sim$0.80   & 96.39$\sim$96.83 & 0.60$\sim$0.80   & 96.14$\sim$96.83 \\ 
OxfordPets    & 0.04$\sim$0.14 & 87.46$\sim$89.07 & 0.04$\sim$0.14 & 88.44$\sim$89.07 \\ 
StanfordCars  & 0.20$\sim$0.30   & 75.35$\sim$76.18 & 0.40$\sim$0.50   & 76.05$\sim$76.18 \\ 
Caltech101     & 0.10$\sim$0.20   & 92.98$\sim$93.83 & 0.10$\sim$0.20   & 93.14$\sim$93.63 \\ 
FGVCAircraft  & 0.60$\sim$0.70   & 45.06$\sim$45.27 & 0.90$\sim$1.00   & 45.15$\sim$45.27 \\ 
SUN397         & 0.06$\sim$0.16 & 70.61$\sim$71.17 & 0.06$\sim$0.16 & 70.33$\sim$71.17 \\ 
UCF101         & 0.40$\sim$0.60   & 79.20$\sim$79.41 & 0.60$\sim$0.90   & 78.98$\sim$79.41 \\ 
\bottomrule
\end{tabular}
\end{adjustbox}
\caption{\textbf{Sensitivity (\%) of Hyperparameters $\beta_2, \beta_3$ on All Datasets.}}
\label{sensitive}
\end{table}

\subsection{Different Visual Encoders}
We experiment CALIP and CALIP-FS with different visual encoders on ImageNet to evaluate their robustness to network architectures. In Table~\ref{encoder}, we list the results of Zero-shot CLIP and CoOp for respectively comparison to CALIP and CALIP-FS. As shown, with different visual networks, CALIP and CALIP-FS still achieve better performance than their baselines.

\begin{table}[t!]
% \vspace{0.3cm}
\small
	\centering
	\begin{adjustbox}{width=0.85\linewidth}
	\begin{tabular}{lccccc}
	\toprule
		\multicolumn{5}{c}{\ \ \ \ \ \ \ \ \ \ \ \ \ \ \ \ \ \ \ \ \ \ \ \ Different Visual Encoders\ \ \ \ \ \ \ \ \ \ \ \ \ \ \ \ \ \ \ \ \ \ \ \ } \\
		\midrule
		\ \ \ Models\ \  &RN50&RN101 &ViT/32 &ViT/16\\
        \cmidrule(lr){1-1} \cmidrule(lr){2-5}
        \specialrule{0em}{1pt}{1pt}
		 \ CLIP &83.94\%&89.17\%&89.57\%&91.40\%\\ 
		 \specialrule{0em}{1pt}{1pt}
        \ CALIP &\textbf{87.51\%}&\textbf{90.67\%}&\textbf{90.91\%}&\textbf{93.14\%}\\
        \cmidrule(lr){1-5}
		 \specialrule{0em}{1pt}{1pt}
        \ CoOp &91.83\%&93.78\%&93.16\%&94.13\%\\ 
		 \specialrule{0em}{1pt}{1pt}
		 CALIP-FS &\textbf{94.20\%}&\textbf{95.13\%}&\textbf{93.23\%}&\textbf{95.74\%}\\ 
		 \specialrule{0em}{1pt}{1pt}
	\bottomrule
	\end{tabular}
\end{adjustbox}
% \vspace*{6pt}
\caption{\textbf{Ablation Study of Different Visual Encoders.}}
% \vspace*{-10pt}
\label{encoder}
\end{table}

\section{Best Dataset Hyperparameters}
We list the best $\beta_1, \beta_2, \beta_3$ for all datasets in Table~\ref{hyper}, where different domains require different proportions of the three logits. For zero-shot CALIP without training, larger $\beta_2$ leads to better recognition accuracy, even surpassing $\beta_1$ for CLIP's original logits on 7 datasets. In contrast, CALIP-FS prefers equal $\beta_2$ and $\beta_3$ and both of them are normally less than $\beta_1$.

\section{Additional Visualization}
We provide more visualizations for cross-modal attention map, spatial visual features and textual features of 16-shot CALIP-FS in Fig~\ref{calip_sup}. As visualized, the attention for texts of the ground-truth category well concentrates on the corresponding objects. After inter-modal interactions, both visual and textual features become more semantic-aligned for the final matching, which fully supports the effectiveness of our approach.

% In Figure~\ref{calip_sup}, we visualize attention maps, spatial visual features before and after the CALIP's non-parametric attention and CALIP-FS's parametric attention, respectively.
% As shown, for both variants, the attention maps concentrate well around the object pixels and the visual features become more distinctive guided by category texts as we expected and illustrate above. We also observe that, after few-shot fine-tuning of CALIP-FS, the distributions of attention maps and visual features all get more intensive, which indicates the improvements resulted from learnable parameters.
\section{Additional Results on Medical Datasets}
To further evaluate the transfer ability of CALIP, we conduct our methods for medical images recognition on two datasets, ISIC~\cite{gutman2016skin} and BCCD~\cite{BCCD}, which contain images of pathological skin injury and blood cells, respectively. The medical images have much more domain gaps than the default eleven 2D datasets and can better test the transfer capacity of CLIP-based models.
We first implement the original CLIP and our CALIP for zero-shot classification. Then, we experiment CALIP-FS for few-shot fine-tuning and also conduct CoOp and Tip-Adapter-F using their officially released codes for comparison.
The results are in Table~\ref{meidical_zs} and Table~\ref{meidical_fs}, where our CALIP and CALIP-FS show leading performance among existing methods.
% \footnote{\url{https://github.com/Shenggan/BCCD_Dataset}}
\vspace{0.2cm}
\begin{table}[h]
% \vspace{0.3cm}
\small
	\centering
	\begin{adjustbox}{width=0.7\linewidth}
	\begin{tabular}{lcccc}
	\toprule
		Datasets  &CLIP &CALIP &Gain\\
        \cmidrule(lr){1-1} \cmidrule(lr){2-2} \cmidrule(lr){3-3} \cmidrule(lr){4-4}
        \specialrule{0em}{1pt}{1pt}
		 ISIC &22.47\%&\textbf{40.62\%}&\textbf{+18.15\%}\\ 
		 \specialrule{0em}{1pt}{1pt}
         BCCD  &36.70\%&\textbf{50.46\%}&\textbf{+13.76\%} \\
	\bottomrule
	\end{tabular}
\end{adjustbox}
% \vspace*{6pt}
\caption{\textbf{Zero-shot Classification on ISIC and BCCD.}}
% \vspace*{-10pt}
\label{meidical_zs}
\end{table}
\vspace{-0.2cm}
\begin{table}[h]
% \vspace{0.3cm}
\small
	\centering
	\begin{adjustbox}{width=0.8\linewidth}
	\begin{tabular}{lcccc}
	\toprule
		Datasets  &CoOp &Tip-Adapter-F &CALIP-FS\\
        \cmidrule(lr){1-1} \cmidrule(lr){2-2} \cmidrule(lr){3-3} \cmidrule(lr){4-4}
        \specialrule{0em}{1pt}{1pt}
		 ISIC &35.66\%&41.46\%&\bf44.54\%\\ 
		 \specialrule{0em}{1pt}{1pt}
         BCCD  &53.21\%&57.80\%&\bf59.63\%\\
	\bottomrule
	\end{tabular}
\end{adjustbox}
% \vspace*{6pt}
\caption{\textbf{16-shot Classification on ISIC and BCCD.}}
% \vspace*{-10pt}
\label{meidical_fs}
\end{table}

\clearpage
% {
% \small
% \bibliographystyle{ieee_fullname}
% % \bibliographystyle{ACM-Reference-Format}
% \bibliography{aaai23}

% \end{document}
%%
%% End of file `sample-sigconf.tex'.

{
\small
\bibliography{aaai23}
}
\clearpage
\end{document}